# Data augmentation for machine learning of chemical process flowsheets

Lukas Schulze Balhorn[a], Edwin Hirtreiter[a], Lynn Luderer[a], Artur M. Schweidtmann[a,*]

[a] *Process Intelligence Research, Department of Chemical Engineering, Delft University of Technology, Van der Maasweg 9, Delft 2629 HZ, The Netherlands*
[*]*Corresponding author. Email: a.schweidtmann@tudelft.nl*

**Abstract**

Artificial intelligence has great potential for accelerating the design and engineering of chemical processes. Recently, we have shown that transformer-based language models can learn to auto-complete chemical process flowsheets using the SFILES 2.0 string notation. Also, we showed that language translation models can be used to translate Process Flow Diagrams (PFDs) into Process and Instrumentation Diagrams (P&IDs). However, artificial intelligence methods require big data and flowsheet data is currently limited. To mitigate this challenge of limited data, we propose a new data augmentation methodology for flowsheet data that is represented in the SFILES 2.0 notation. We show that the proposed data augmentation improves the performance of artificial intelligence-based process design models. In our case study flowsheet data augmentation improved the prediction uncertainty of the flowsheet autocompletion model by 14.7%. In the future, our flowsheet data augmentation can be used for other machine learning algorithms on chemical process flowsheets that are based on SFILES notation.

## 1. Introduction

The design of a flowsheet topology is an important step in early process synthesis. This step consists of selecting and arranging unit operations for a chemical process. Artificial intelligence (AI) methods have the potential to learn from previous flowsheets and support engineers in process development (Hirtreiter et al., 2022; Oeing et al., 2022; Schweidtmann, 2022; Vogel et al., 2023). For instance, Vogel et al. (2023) proposed an algorithm for the autocompletion of flowsheets. This autocompletion algorithm is inspired by text-autocompletion from natural language processing (NLP) that is based on generative transformer models (Radford et al., 2019). In addition, Hirtreiter et al. (2022) showed that the prediction of control structure elements from Process Flow Diagrams (PFDs) can be interpreted as a translation task between PFDs and Process and Instrumentation Diagrams (P&IDs). Hence, they deployed a sequence-to-sequence transformer architecture which is commonly used for translation of text between different languages. These flowsheet transformers rely on machine-readable flowsheet representations.

To represent flowsheets in a machine-readable format, we depict them as graphs or as text using unique, i.e., canonical, SFILES 2.0 strings (Vogel et al., 2022b). In general, flowsheets are drawings of chemical processes. Chemical engineers use flowsheets for the communication, planning, operation, simulation, and construction of these processes. An example flowsheet is given in Figure 1.

An intuitive way to represent flowsheets is via graphs with unit operations as nodes and stream connections as directed edges. Besides the graph representation, flowsheets can also be represented as strings. D'Anterroches (2005) introduced the Simplified Flowsheet Input-Line Entry-System (SFILES) notation, which we recently extended to include control structures and other features in (Vogel et al., 2022b). When creating the SFILES, we traverse the graph by starting at an input node and following the stream direction until we reach a product node or a recycle. In case the stream branches at a node, i.e., a splitter, we need to decide which stream to follow first. To determine the order of the branches in the linear string, the SFILES algorithm assigns each node a unique rank. The SFILES string for the flowsheet from Figure 1 is given by:

(raw)(hex){1}(r)<&|(raw)(pp)&|(mix)<1(v)(dist)[{tout}(prod)]{bout}(splt)1(prod)n|(raw)(hex){1}(prod).

While AI models require big training data, machine-readable flowsheet data is typically limited. The reason for limited data is that flowsheets are mainly depicted as images and therefore not machine-readable (Schweidtmann, 2022). Recently, we propose to automatically find flowsheet images in literature (Schulze Balhorn et al., 2022) and to make them machine-readable via computer vision (Theisen et al., 2023). However, this methods need to be implemented at a large scale. In addition, the majority of flowsheets are not publicly available due to company's intellectual property protection. Chemical process datasets with machine-readable flowsheets are rare and often contain only dozens of flowsheets (Hirtreiter et al., 2022; Oeing et al., 2022; Vogel et al., 2023; Zhang et al., 2018; Zheng et al., 2022). However, AI methods like transformers usually require big training data.

One promising approach to overcome limited training data for artificial intelligence is data augmentation. Data augmentation builds on the idea to generate additional artificial training data by masking, modifying, perturbating the available original training data at hand. For example, in computer vision, data augmentation is an established method to improve the model performance by adding modified copies of already existing images (Shorten & Khoshgoftaar, 2019). However, no data augmentation method for flowsheets exists yet. Thus, a augmentation method is needed which adds modified copies of already existing flowsheets to mitigate the issue of limited available flowsheet data.

We propose a novel data augmentation method for process flowsheets. Specifically, we use a text-based augmentation method for SFILES which is inspired by the augmentation of SMILES strings for molecules (Bjerrum, 2017). Our approach is to randomize the branching decisions in the SFILES generation. We demonstrate the proposed flowsheet augmentation method in the context of a flowsheet autocompletion model from (Vogel et al., 2023).

## 2. Data augmentation methodology

To augment the flowsheet data sets, we modify the branching decision in the SFILES generation algorithm to create multiple SFILES strings representing the same flowsheet. In the case of determining canonical SFILES the branching decisions are made by assigning every node of the graph to a unique rank. Hence, canonical SFILES are a unique mapping of a flowsheet graph to a string. When generating augmented SFILES, the branching decisions are made randomly, resulting in a non-canonical form. The difference lies in the order of branches in the linear string. The resulting augmented SFILES contain the identical information as the canonical SFILES, thus describing the same process flowsheet. Hence, all augmentations can be translated back to the original canonical SFILES. In Figure 1 we show the augmentation of the example flowsheet.

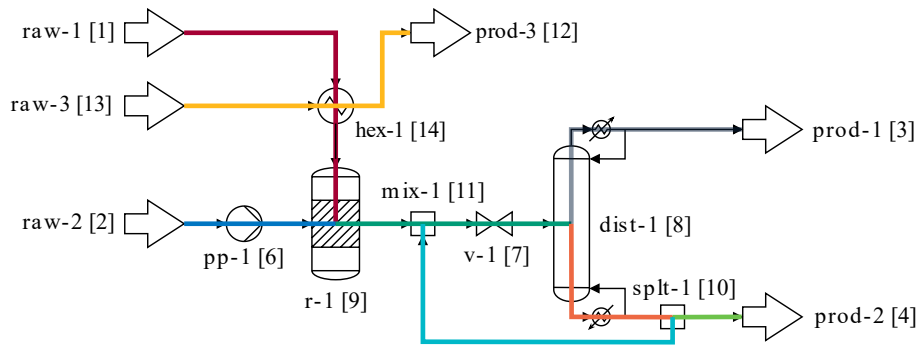

Figure 1: Augmentation example for the flowsheet from Figure 1. The node rank is given in square brackets after the node name.

During augmentation, only the uniqueness of the SFILES representation is lost while the full flowsheet topology information is preserved. Notably, we do not change the order of the input branches during data augmentation. Otherwise, flowsheets with disconnected sub-graphs cannot be translated back to the original canonical form. This is for example important for independent processes with heat integration, such as *(raw)(hex){1}(prod)* in the example flowsheet from Figure 1.

The number of augmented SFILES that can be derived from a single flowsheet graph is limited. Specifically, the number of potential augmented SFILES depends on the number of available branching points for a given flowsheet, with more branching points leading to an exponential increase of augmentation possibilities. For example, the flowsheet in Figure 1 contains two branching points. This results in three augmented SFILES representations and four SFILES representations in total.

The two branching points in this case are the distillation column *dist-1* and the splitter after the distillation column *splt-1*. At the first branching point, we switch the order of the top and bottom outlet streams. The second augmentation affects only the bottom product stream *prod-2*. Recycles without unit operations always appear directly after the splitting unit, here that leads to *(splt-1)1* in all cases. In case the product branch is visited before the recycle branch, additional squared brackets around *prod-2* are used. To make the methodology for flowsheet augmentation openly accessible, we include it in our public SFILES 2.0 Github repository (Vogel et al., 2022a).

## 3. Case study and Results

### 3.1. Data and Data augmentation

We use two SFILES datasets which were created by (Vogel et al., 2023). Firstly, we use their proposed flowsheet generator to generate a large-scale dataset of about 8,000 artificial flowsheets. Here, we can flexibly scale the dataset size. Secondly, we use a dataset made from 223 Aspen and DWSIM chemical process simulations. We call this the real dataset. Before starting the training runs, we split each dataset into a training dataset, a validation dataset, and a test dataset. For the two training datasets we created a maximum of five augmentations for each SFILES, which roughly increases the dataset five-fold. By limiting the number of augmentations to five, we ensure that larger flowsheets are not over-represented.

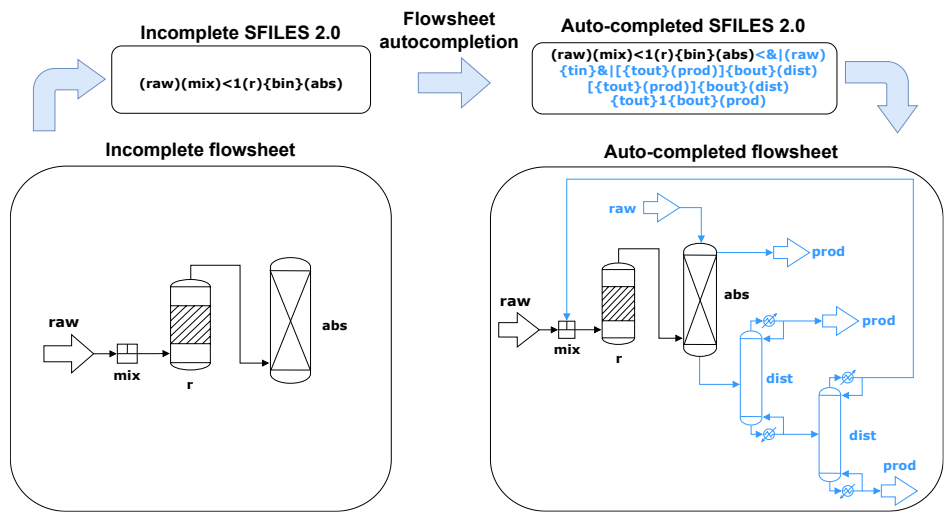

Figure 2: Example prediction of the flowsheet autocompletion model. Figure adapted from (Vogel et al., 2023).

### 3.2. Flowsheet autocompletion

We consider a process design case study and use the flowsheet autocompletion model from (Vogel et al., 2023). The objective of the model is to support chemical engineers in the design of a new process topology. It can suggest the following unit operation for an incomplete flowsheet, similar to sentence completion in messenger apps. Specifically, the start of a SFILES is given and the autocompletion model predicts how the sequence ends by iteratively predicting the following building block, token, of the SFILES (Figure 2). It should be highlighted that the prediction is only based on the process topology and does not consider operating points, components, and material flows. The autocompletion model is built on the transformer architecture in a decoder-only approach (Radford et al., 2019). Because transformer models are very data-intensive, the generated dataset is used to pretrain the flowsheet autocompletion. The real dataset is then used to fine tune the flowsheet autocompletion. For a more detailed description of the model we refer to (Vogel et al., 2023).

To study the effectiveness of data augmentation we retrain the model for flowsheet autocompletion from (Vogel et al., 2023). Specifically, we train the model both with augmented and non-augmented flowsheet data. Overall, we consider three different training scenarios. First, we train the model with the non-augmented generated dataset and fine-tune this model with the real dataset (i). This model is used to reproduce the results from (Vogel et al., 2023). Secondly, we augment the real data and use them to fine-tune the pretraining model, resulting in model (ii). We thus only alter the fine-tuning. Finally, we train the flowsheet autocompletion with augmented data for both pretraining and fine-tuning, yielding model (iii). For a fair comparison of training runs with and without data augmentation, we use the same hyperparameters.

*3.3. Results*

We use perplexity to measure the model performance. Perplexity describes the uncertainty of a model in its predictions. Therefore, a low perplexity is desirable. Here, a lower perplexity means that the model is more confident that the suggested unit operation is reasonable. Perplexity is the exponential of the negative average log-likelihood of the next token prediction. It is also equivalent to the exponential of the cross-entropy loss obtained during model training. The perplexity is computed as

$$PP(T) = \exp\left(-\frac{1}{n}\sum_{i}^{n} \log P(t_i|t_{1:i-1})\right), \quad (1)$$

where $T = (t_1, \ldots, t_n)$ is a sequence of $n$ tokens and $P(t_i|t_{1:i-1})$ describes the predicted probability for the next token $t_i$ given the sequence $t_{1:i-1}$.

Table 1: Perplexity $PP$ results of data augmentation. Best test result in bold font. The column "Augm." shows whether the training data were augmented or not. Pretraining and fine-tuning test perplexity are both evaluated after the model training is completed with the fine-tuning dataset.

| Model | $PP$ pretraining | | $PP$ fine-tuning | | | | Training Time |
|---|---|---|---|---|---|---|---|
| | Augm. | Test | Augm. | Train | Val | Test | |
| (i) | No | 5.38 | No | 3.13 | 4.23 | 5.02 | 51 min |
| (ii) | No | 6.33 | Yes | 3.32 | 4.12 | 4.69 | 57 min |
| (iii) | Yes | **5.16** | Yes | 3.07 | 3.80 | **4.28** | 1 h 31 min |

The results of the different training runs are shown in Table 1. For the fine-tuning, model (ii) performs slightly better than model (i) on the test set. We explain this improvement by the fact that the fine-tuning data are limited in size (i) and with the data augmentation we can make the model more robust (ii). Even though the test perplexity is lower with data augmentation, the training perplexity is higher. This shows that the model trained with data augmentation is less prone to overfitting and generalizes better to unseen data. With data augmentation also applied to the pretraining (model (iii)), we see the best performance in all categories, improving the fine-tuning perplexity by 14.7% compared to model (i). We conclude that data augmentation can also improve the pretraining with a relatively large, generated dataset, resulting in a better fine-tuning performance.

In general, we see that the augmented SFILES are valid flowsheet representations and that they help the flowsheet autocompletion to learn the SFILES grammar and chemical process structure. For future work it would be interesting to investigate if it is more favorable for pretraining to generate more artificial data, to increase the dataset with data augmentation, or to combine both methods. It is worth noticing, that the training and test perplexity are in every case more similar for the models trained with augmented data. Due to the higher variance in the training data, these models are less prone to overfitting.

## 4. Conclusions

We propose a data augmentation of chemical flowsheet data by randomizing the branching decisions in the graph traversal for producing SFILES. We thereby demonstrate a way to increase the available flowsheet data for subsequent training of AI models. We apply augmented SFILES to the problem of flowsheet autocompletion and show that the augmentation improves the model performance in low data regimes. In future research, we aim to apply the data augmentation methodology to further NLP models to improve their performance. The long-term goal should be to increase the flowsheet dataset size by mining additional flowsheets from literature (Schweidtmann, 2022; Schulze Balhorn et al., 2022) and companies and, if necessary, digitizing them (Theisen et al., 2023). A combination of both, data augmentation and increased dataset, size will be necessary and most beneficial.

**References**

Bjerrum, E. J. (2017). Smiles enumeration as data augmentation for neural network modeling of molecules. arXiv preprint arXiv:1703.07076.

d'Anterroches, L. (2005). Process flowsheet generation & design through a group contribution approach. [CAPEC], Department of Chemical Engineering, Technical University of Denmark.

Hirtreiter, E., Schulze Balhorn, L., & Schweidtmann, A. M. (2022). Towards automatic generation of Piping and Instrumentation Diagrams (P&IDs) with Artificial Intelligence. arXiv preprint arXiv:2211.05583.

Oeing, J., Welscher, W., Krink, N., Jansen, L., Henke, F., & Kockmann, N. (2022). Using artificial intelligence to support the drawing of piping and instrumentation diagrams using dexpi standard. Digital Chemical Engineering, 4, 100038.

Radford, A., Wu, J., Child, R., Luan, D., Amodei, D., & Sutskever, I. (2019). Language models are unsupervised multitask learners. OpenAI blog, 1(8), 9.

Schulze Balhorn, L., Gao, Q., Goldstein, D., & Schweidtmanna, A. M. (2022). Flowsheet Recognition using Deep Convolutional Neural Networks. In Computer Aided Chemical Engineering (Vol. 49, pp. 1567-1572). Elsevier.

Schweidtmann, A. M. (2022). Flowsheet mining. Manuscript, In preparation. TU Delft.

Shorten, C., & Khoshgoftaar, T. M. (2019). A survey on image data augmentation for deep learning. Journal of big data, 6(1), 1-48.

Theisen, M. F., Flores, K. N., Schulze Balhorn, L., & Schweidtmann, A. M. (2023). Digitization of chemical process flow diagrams using deep convolutional neural networks. Digital Chemical Engineering, 6, 100072.

Vogel, G., Schulze Balhorn, L., Hirtreiter, E., & Schweidtmann, A. M. (2022a). Process-intelligence-research/sfiles2: V1.0.0 (Version Release). Github. https://github.com/process-intelligence-research/SFILES2

Vogel, G., Schulze Balhorn, L., Hirtreiter, E., & Schweidtmann, A. M. (2022b). SFILES 2.0: An extended text-based flowsheet representation. arXiv preprint arXiv:2208.00778.

Vogel, G., Schulze Balhorn, L., & Schweidtmann, A. M. (2023). Learning from flowsheets: A generative transformer model for autocompletion of flowsheets. Computers & Chemical Engineering, 171, 108162.

Zhang, T., Sahinidis, N. V., & Siirola, J. J. (2019). Pattern recognition in chemical process flowsheets. AIChE Journal, 65(2), 592-603.

Zheng, C., Chen, X., Zhang, T., Sahinidis, N. V., & Siirola, J. J. (2022). Learning process patterns via multiple sequence alignment. Computers & Chemical Engineering, 159, 107676.